\documentclass{article}
\usepackage{spconf,amsmath,graphicx}
\usepackage{algorithm2e}
\usepackage{hyperref}
\usepackage{verbatim}
\usepackage{subcaption}
\usepackage{amssymb}
\usepackage{multirow}
\usepackage{xcolor}
\usepackage{enumitem}
\usepackage{dcolumn}
\usepackage{threeparttable}
\usepackage{booktabs}
\usepackage{url,footnote}
\usepackage{breqn}
\usepackage{here}
\DeclareUnicodeCharacter{0002}{fi}
\let\emph\textit

\title{Non-autoregressive Transformer with Unified Bidirectional Decoder for Automatic Speech Recognition}

\name{
    \begin{tabular}{c}
       \large ~Chuan-Fei Zhang$^{1, 2}$,Yan Liu$^{1, 2, *}$\thanks{* Corresponding author.},~Tian-Hao Zhang$^{3}$,~Song-Lu Chen$^{3}$,~Feng Chen$^{3,4}$,~Xu-Cheng Yin$^{3}$
   \end{tabular}
    \vspace{-5pt}
}
\address{
    $^1$\normalsize School of Automation and Electrical Engineering, University of Science and Technology Beijing, Beijing 100083, China\\
    $^2$ \normalsize Key Laboratory of Knowledge Automation for Industrial Processes of Ministry of Education, Beijing 100083, China\\
    $^3$ \normalsize USTB-EEasyTech Joint Lab of Artificial Intelligence, University of Science and Technology Beijing, Beijing 100083, China\\
    $^4$ \normalsize EEasy Technology Company Ltd., Zhuhai 519000, China
	\vspace{-10pt}
}

\begin{document}
\ninept

\maketitle

\begin{abstract}
Non-autoregressive (NAR) transformer models have been studied intensively in automatic speech recognition (ASR), and a substantial part of NAR transformer models is to use the casual mask to limit token dependencies. However, the casual mask is designed for the left-to-right decoding process of the non-parallel autoregressive (AR) transformer, which is inappropriate for the parallel NAR transformer since it ignores the right-to-left contexts. Some models are proposed to utilize right-to-left contexts with an extra decoder, but these methods increase the model complexity. To tackle the above problems, we propose a new non-autoregressive transformer with a unified bidirectional decoder (NAT-UBD), which can simultaneously utilize left-to-right and right-to-left contexts. However, direct use of bidirectional contexts will cause information leakage, which means the decoder output can be affected by the character information
from the input of the same position. To avoid information leakage, we propose a novel attention mask and modify vanilla queries, keys, and values matrices for NAT-UBD. Experimental results verify that NAT-UBD can achieve character error rates (CERs) of 5.0\%/5.5\% on the Aishell1 dev/test sets, outperforming all previous NAR transformer models. Moreover, NAT-UBD can run 49.8$\times$ faster than the AR transformer baseline when decoding in a single step.

\end{abstract}


\section{Introduction}
\label{sec:intro}

Recently, transformer models \cite{Linhao2018speech,Yingzhu2020Persistent} based on encoder-decoder have shown superior performance in end-to-end automatic speech recognition (ASR) compared with Recurrent Neural Networks (RNNs) \cite{Chiu2020state,Li2018Fast} and Connectionist Temporal Classification (CTC) \cite{alex2006ctc}. Deficiently, most transformer models predict the next token conditioning on encoded states and previously generated tokens in an autoregressive (AR) manner, resulting in slow decoding speed.

\begin{figure} [t]
     \centering
     \includegraphics[scale=0.65]{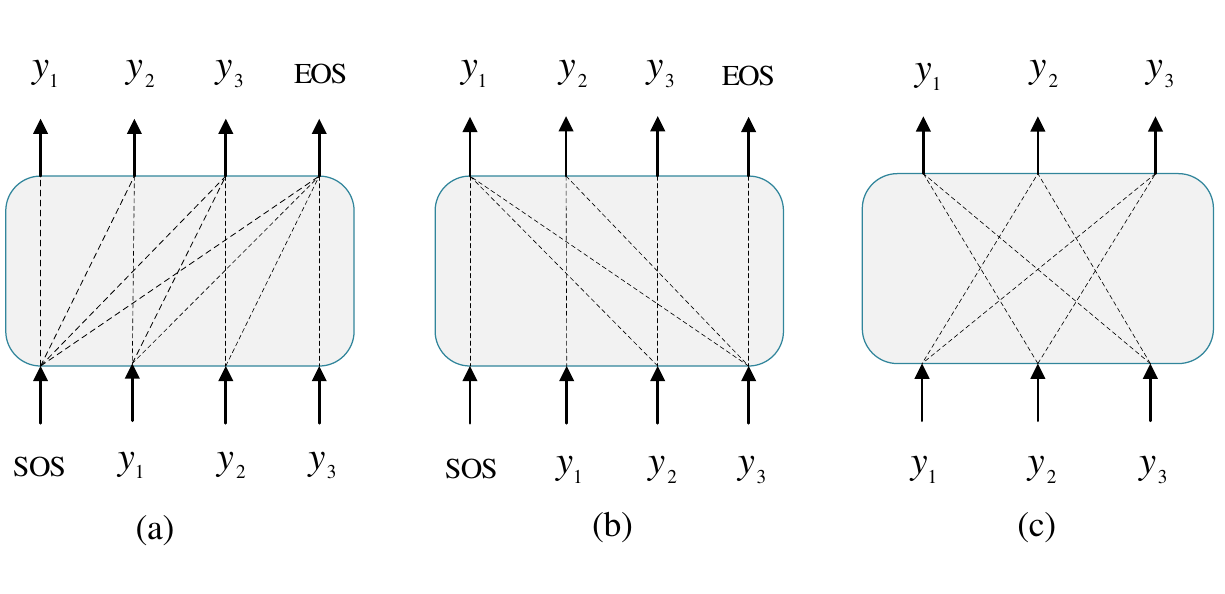}
     \caption{(a) unidirectional decoder \cite{song2021duokuileni,TSNAT2021TWOPASS} with left-to-right contexts only; (b) unidirectional decoder \cite{bidireation2020AR,bidirectional2021stream} with right-to-left contexts only; (c) unified bidirectional decoder with right-to-left contexts and left-to-right contexts simultaneously. The grey boxes denote transformer decoder, and the dashed lines denote token dependencies.}
\label{fig.intro}
\end{figure}

To accelerate the decoding speed, non-autoregressive (NAR) transformer models \cite{gu2018trans,MARJAN2020ENTROPY,JASON2018DETERMINISTIC} are first proposed in machine translation, which can predict multiple tokens simultaneously and have been widely studied
in ASR recently. To our best knowledge, NAR transformer models in ASR can be roughly divided into two categories according to the decoder. The first kind of NAR transformer model \cite{zhengkun2020spike,cass2020nat} regards the decoder as an acoustic model. However, such NAR transformer models follow the conditional independence hypothesis between the output tokens and suffer inferior recognition performance. The second kind of NAR transformer model \cite{Yosuke2020maskctc,YEBAI2020laji,song2021duokuileni,TSNAT2021TWOPASS} regards the decoder as a language model, and the decoder can predict output conditioning on linguistic information. Notably, the attention mask is widely used in these NAR transformers to limit token dependencies. Especially, the casual mask proposed in the AR transformer \cite{attention} is used in the second kind of NAR transformers \cite{song2021duokuileni,TSNAT2021TWOPASS} to construct a unidirectional decoder (Fig.~\ref{fig.intro} (a)). However, it is inappropriate for NAR transformer models to use the casual mask. Firstly, the casual mask is designed for the serial decoding process of the AR transformer while the decoding process of the NAR transformer is parallel. Secondly, the casual mask only uses left-to-right (L2R) contexts, resulting in discarded right-to-left (R2L) contexts.

Previously, R2L contexts (Fig.~\ref{fig.intro} (b)) have been studied in the AR transformer \cite{bidireation2020AR} and the streaming ASR \cite{bidirectional2021stream}. These models are composed of one shared encoder and two unidirectional decoders, i.e., two separate decoders with L2R and R2L contexts, respectively. Such a framework is complex and inefficient because it needs an extra unidirectional decoder and the two decoders have no information exchange.

To tackle the above problems, we propose a new non-autoregre-
ssive transformer with a unified bidirectional decoder (NAT-UBD), which can fully utilize both L2R and R2L contexts in a unified decoder (Fig.~\ref{fig.intro} (c)). However, direct use of bidirectional contexts will cause information leakage. Concretely, information leakage means the decoder output can be affected by the character information from the input of the same position, and the decoder can not refine the input during decoding. Since the proposed NAT-UBD is based on the speech transformer \cite{Linhao2018speech}, the residual connection and self-attention mechanism are adopted, and both of them can cause information leakage. To avoid information leakage resulting from the residual connection, we remove word embedding from vanilla queries matrix ($Q$). To avoid information leakage resulting from the self-attention mechanism, we propose a novel attention mask named self mask and make both keys matrix ($K$) and values matirx ($V$) independent of layers, similar to the Disco transformer \cite{GUJIATAO2020dashen}. This way, NAT-UBD can outperform all previous NAR transformer models on the Aishell1 corpus and achieve competitive performance compared with the AR transformer baseline on the Magicdata corpus. Moreover, NAT-UBD can run much faster than the AR transformer baseline because UBD can predict all tokens simultaneously.

\section{Methodology}
\label{sec:Non-autoregressive transformer with bidirectional contexts}

\begin{figure} [t]
     \centering
     \includegraphics[scale=0.54]{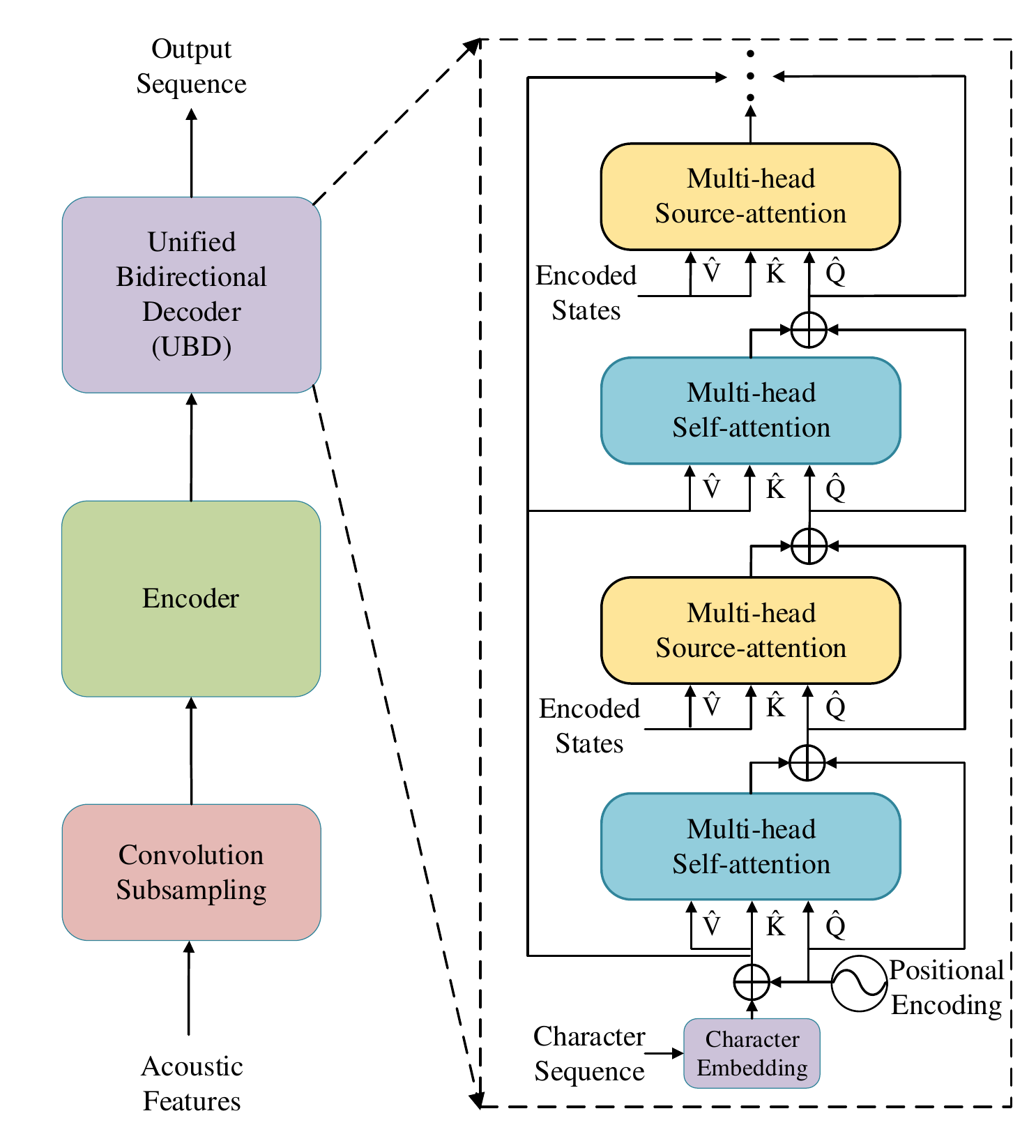}
     \caption{The overall architecture of NAT-UBD. First, the input acoustic features are downsampled by convolutional subsampling layers and encoded by the encoder. Then the encoded states and character sequence are fed into UBD. Finally, UBD predicts all tokens simultaneously.}
\label{fig.model}
\end{figure}

The proposed NAT-UBD can fully utilize both L2R and R2L contexts in a unified decoder. Fig.~\ref{fig.model} illustrates the overall architecture of NAT-UBD. For simplicity, we omit the feed-forward layers and layer normalization \cite{layernormalization}. The convolutional
subsampling layers and encoder of NAT-UBD are the same as the speech transformer \cite{Linhao2018speech}, while the decoder is our proposed unified bidirectional decoder (UBD).
\subsection{Unified Bidirectional Decoder}
\label{sec:Decoder_objective}

UBD takes the character sequence $Y$ as the input and predicts all tokens simultaneously conditioning on bidirectional contexts $Y_{\neq t}$ and encoded states $S$, as described in Eq~(\ref{eq:forward}).

\begin{equation}
    \label{eq:forward}
    y_t = UBD({Y}_{\neq t}, S)   \qquad 1 \le t \le T,
\end{equation}
where $T$ is the length of $Y$. Besides, $Y_{\neq t}$ represents that the output $y_t$ should prevent from utilizing the character information from the input of the same position. Otherwise, information leakage will arise.

In the vanilla transformer, the character sequence $Y$ is first transformed to $QKVs$ by the character embedding $C$ and positional encoding $P$, as described in Eq~(\ref{eq:VanillaQ}).

\begin{equation}
\label{eq:VanillaQ}
Q^1, K^1, V^1 = Linear(C(Y)+P)
\end{equation}

\begin{figure} [t]
     \centering
     \includegraphics[scale=0.54]{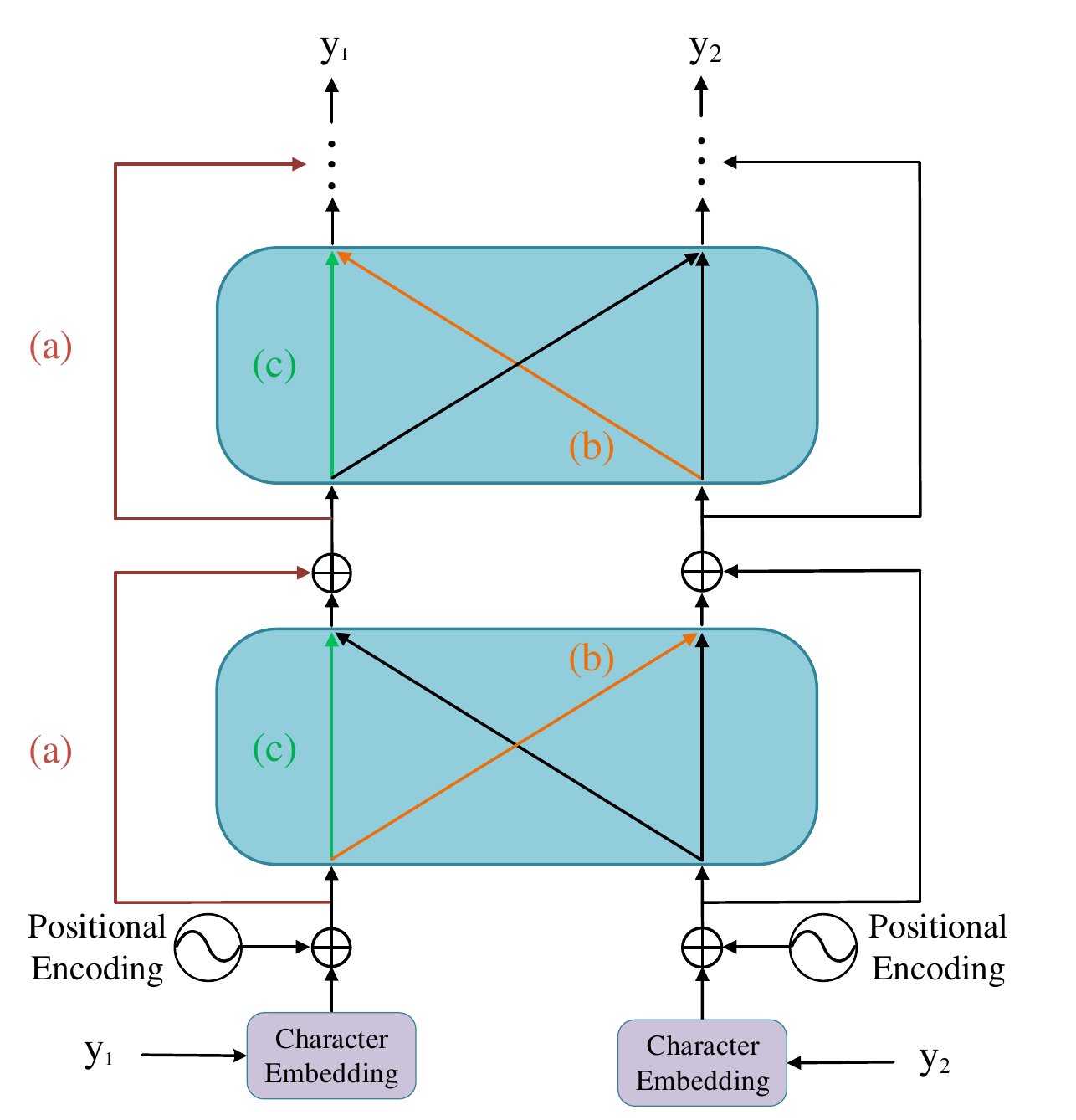}
     \caption{Illustration of information leakage. Both residual connections and multi-head self-attention layers can cause information leakage. (a) residual connections; (b) indirect attention connections; (c) direct attention connections. Blue boxes denote multi-head self-attention layers.
    }
\label{fig.informationleakage}
\end{figure}

For simplicity, we use $Linear()$ to represent different linear layers. This way, $QKVs$ of the first multi-head
self-attention layer contain the character information and are transmitted by residual connections and multi-head self-attention layers. However, both residual connections (Fig.~\ref{fig.informationleakage} (a)) and multi-head self-attention layers (Fig.~\ref{fig.informationleakage} (b)(c)) can cause information leakage.
% We omit linear layers and multi-head source-attention layers for simplicity.

To avoid information leakage caused by residual connections, all queries of $Q$ should not contain the character information of the current position. So we remove the character embedding and calculate the queries matrix of the first multi-head self-attention layer, as described in Eq~(\ref{eq:Q1}).

\begin{equation}
    \label{eq:Q1}
    \hat{Q}^{1} = Linear(P),
\end{equation}
where $\hat{Q}$ represents the queries matrix of NAT-UBD. However, $KVs$ can not be modified like $\hat{Q}$ since the character information must be retained and utilized in UBD. An alternative method is to remove residual connections for $KVs$. Nevertheless, removing residual connections is not enough since the attention connections in multi-head self-attention layers will also cause information leakage.

To avoid the information leakage caused by indirect attention connections (Fig.~\ref{fig.informationleakage} (b)), all keys and values in multi-head self-attention layers should only contain the character information of the current position. So we feed the same keys and values matrics independent of the previous layers into all multi-head self-attention layers, as described in Eq~(\ref{eq:KV}).
%, to prevent all positions from paying attention to themselves cyclicly:

\begin{equation}
    \label{eq:KV}
    \hat{K}^i, \hat{V}^i = Linear(C(Y)+P)   \quad  1 \le i \le I,
\end{equation}

where $\hat{K}$ and $\hat{V}$ are keys matrix and values matrix of NAT-UBD. Besides, $I$ is the number of multi-head self-attention layers.

To avoid the information leakage caused by direct attention connections (Fig.~\ref{fig.informationleakage}), the diagonal elements of the attention weight matrix should be set to 0. Hence, we propose an attention mask named self mask to multiply with the attention weight matrix. Concretely, the self mask is an attention mask matrix of which diagonal elements are 0, and all the other elements are 1. In addition, the self mask also makes UBD can utilize bidirectional contexts simultaneously.

\subsection{Joint Training}
To enable the encoder to output preliminary results for UBD during decoding, we apply the CTC loss \cite{alex2006ctc} to the encoder. Besides, we use the label sequence as the input character sequence of UBD. Then NAT-UBD can be jointly trained with both the CTC loss $L_{CTC}$ and UBD loss $L_{UBD}$, as described in Eq~(\ref{eq:training_loss}).
\begin{equation}
    \label{eq:training_loss}
    L = \lambda L_{CTC}+(1-\lambda)L_{UBD},
\end{equation}
where $\lambda$ is a hyperparameter to balance two losses. $L_{UBD}$ is the cross entropy loss \cite{CE}.
\subsection{Decoding by Iterative Refinement and Early Termination}
\label{sec:Decoding}

To achieve fast decoding speed, UBD takes the greedy CTC outputs to substitute the label sequence as the initial input. All decoder inputs can use bidirectional contexts to predict, and the whole output sequence can be predicted by iterative refinement in $J$ iterations, as described in Eq~(\ref{eq:decoding}).
\begin{equation}
\label{eq:decoding}
\hat{y}_t^j=
\left\{
\begin{array}{lr}
UBD(\hat{Y}_{\neq t}^{j-1,CTC},S)  &j=1, \\
UBD(\hat{Y}_{\neq t}^{j-1,UBD},S)  &1<j \le J,
\end{array}
\right.
\end{equation}
where $\hat{Y}^{j-1,CTC}$ is the greedy CTC outputs, and $\hat{Y}^{j-1,UBD}$ is the greedy UBD outputs.

Moreover, we propose a simple but effective stop method named adaptive termination for NAT-UBD to accelerate decoding speed. When the output of the $j$th iteration is the same as the $(j-1)$th iteration, the iteration can be early terminated because the output of subsequent iteration will remain unchanged.

\section{Experiments}
\label{sec:Experiments}

\subsection{Datasets}

The experiments are carried out on the 178-hour  Aishell1 \cite{buhui2017aishell1} Mandarin corpus and 755-hour Magicdata $\footnotemark[1]$ Mandarin corpus. For the input acoustic features, we extract $80$-channel filterbanks features splice $3$-channel pitch computed from a $25$ms window with a stride of $10$ms. The output labels consist of 4231 Chinese characters for Aishell1 and 4518 Chinese characters for Magicdata, obtained from the training set.

\footnotetext[1]{~~Beijing Magic Data Co., Ltd. www.magicdatatech.com/}

\subsection{Experimental Setup}

We use ESPNet \cite{taaki2018tespnet} for all experiments. The convolutional subsampling module is comprised of 2 CNN layers with size 3×3, filter 256, stride 2 on the time dimension for 4× down-sampling. Then 12 encoder layers and 6 decoder layers are stacked. We use 256 dimensions for $\hat{Q}\hat{K}\hat{V}s$  and 4 attention heads for all multi-head attention layers. We set 2048 dimensions for the position-wise feed-forward networks and use ReLU activation for the hidden layer. Label smoothing \cite{labelsmoth} with a penalty of 0.1 is applied to prevent over-fitting. We use Adam \cite{adam} optimizer and warm up with $\beta_{1}=0.9$, $\beta_{2}=0.98$, and $\epsilon=10^{-9}$. All models are trained on 2 Titan X GPUs with batch size 32. Gradients are accumulated \cite{MYLE2018scaling} over 4 iterations. SpecAugment \cite{PARK2019specaugment} is used for data augmentation, and Speed Perturbation \cite{ko2015speed} is additionally used for the Aishell1 corpus. We use the dev set for early stopping. We choose the best models of 10 epochs with the lowest accuracies on the dev set and average them to get the final model. All decoding processes are performed utterance by utterance on a Titan X GPU without any external language model. Character error rate (CER), character error rate reduction (CERR), and real time factor (RTF) are adopted for model evaluation. RTF is computed as the ratio of the total decoding time to the total duration of the test set.

\subsection{Results}

\begin{table}[t]
\caption{Character error rate (CER) and real time factor (RTF) on the Aishell1 corpus. \dag \ means SpecAugment is used, and \ddag \ means Speed Perturbation is used. $J$ is the maximum number of iterations, and $T$ is the length of the output sequence.}
\label{tab:aishell}
\centering
\scalebox{0.95}{
\begin{tabular}{cccccc}
\hline
Model & $J$ & Dev & Test & RTF & Speedup \\
\hline
\multicolumn{6}{c}{\emph{AR transformer}} \\
Transformer ~\cite{taaki2018tespnet}\ddag & $T$ &6.0 & 6.7 & $-$ & $-$ \\
\hline
\multicolumn{6}{c}{\emph{NAR transformer}} \\
LASO-small ~\cite{YEBAI2020laji}\dag\ddag & 1 & 6.0 & 6.8 & $-$ & $-$ \\
ST-NAR ~\cite{zhengkun2020spike}\ddag & 1 & 6.9 & 7.7 & $-$ & $-$ \\
KERMIT ~\cite{fujita2020insertion}\ddag & 1 & 6.7 & 7.5 & $-$ & $-$ \\
InDIGO ~\cite{fujita2020insertion}\ddag & 1 & 6.0 & 6.7 & $-$ & $-$ \\
CASS-NAT ~\cite{cass2020nat}\dag\ddag & 1 & 5.3 & 5.8 & $-$ & $-$ \\
CTC-enhanced ~\cite{song2021duokuileni}\dag\ddag & 1 & 5.3 & 5.9 & $-$ & $-$\\
A-FMLM ~\cite{chen2020nanshou}\ddag & 1 & 6.2 & 6.7 & $-$ & $-$ \\
TSNAT-small ~\cite{TSNAT2021TWOPASS}\ddag & 1 & 5.4 & 5.9 & $-$ & $-$ \\
\hline
\multicolumn{6}{c}{\emph{Our work}} \\
AR transformer\dag\ddag & $T$ & 5.2 & 5.6 & 0.4034 & 1.00$\times$ \\
NAT-UBD\dag\ddag & 1 & 5.1 & 5.6 & \textbf{0.0081} & \textbf{49.8}$\times$ \\
NAT-UBD\dag\ddag & 10 & \textbf{5.0} & \textbf{5.5} & 0.0116 & 34.8$\times$ \\
\hline
\end{tabular}}
\end{table}

\begin{table}[t]
\caption{Character error rate (CER) and real time factor (RTF) on the Magicdata corpus. \dag \ means SpecAugment is used.}
\label{tab:magicdata}
\centering
\scalebox{0.95}{
\begin{tabular}{cccccc}
\hline
Model & $J$ & Dev & Test & RTF & Speedup \\
\hline
\multicolumn{6}{c}{\emph{Our work}} \\
AR transformer\dag & $T$ & \textbf{4.1} & \textbf{4.6} & 0.4703 & 1.00$\times$ \\
NAT-UBD\dag & 1 & 4.4 & 4.9 & \textbf{0.0095} & \textbf{49.5}$\times$ \\
NAT-UBD\dag & 10 & 4.2 & 4.7 & 0.0121 & 38.9$\times$ \\
\hline
\end{tabular}}
\end{table}
Except for NAT-UBD, we reimplement the AR transformer baseline with CTC joint training and joint decoding \cite{shigeki2020imrpoving}. Especially, beam search with a width of 10 is used for the AR transformer baseline. As shown in Table ~\ref{tab:aishell}, our proposed NAT-UBD achieves the best CERs than all previous NAR transformer models with different iterations and can outperform the AR transformer baseline with a 49.8$\times$ faster decoding speed on the Aishell1 corpus.

We conduct experiments on the Magicdata corpus. Magicdata is a large Mandarin corpus, and to our best knowledge, we are the first to assess the NAR transformer models on this corpus. From Table ~\ref{tab:magicdata}, we can see that NAT-UBD can achieve competitive CERs with the AR transformer baseline while maintaining faster decoding speed, proving the generalizability of NAT-UBD.

\subsection{Necessity of Avoiding Information Leakage}

\begin{table}[t]
\caption{Ablation study of $\hat{Q}\hat{K}\hat{V}s$ and self mask on the Aishell1 corpus. $J = 0$ means that the greedy CTC output is directly used as the final output.}
\label{tab:leakage}
\centering
\scalebox{0.95}{
\begin{tabular}{lccccc}
\hline
\multicolumn{1}{c}{\multirow{2}{*}{Model}} & {\multirow{2}{*}{$J$}} & \multicolumn{2}{c}{Dev} & \multicolumn{2}{c}{Test}\\ \cmidrule(r){3-4} \cmidrule(r){5-6}
\multicolumn{1}{c}{} & & CER & CERR& CER & CERR           \\
\hline
\multirow{2}{*} {NAT-UBD} & 0 & 5.5& $-$ & 6.0& $-$  \\
& 10 & \textbf{5.0} & \textbf{9.1} & \textbf{5.5} & \textbf{8.3}  \\
%\hline
\ \multirow{2}{*} {$-$$\hat{Q}\hat{K}\hat{V}s$} & 0 & 12.7 & $-$ & 13.6 & $-$  \\
& 10 & 12.7 & 0.0 & 13.6 & 0.0  \\
%\hline
\ \multirow{2}{*} {$-$self mask} & 0 & 11.1 & $-$ & 12.6 & $-$  \\
& 10 & 11.1 & 0.0 & 12.6 & 0.0  \\
\ \multirow{2}{*} {$-$both} & 0 & 10.6 & $-$ & 11.3 & $-$  \\
& 10 & 10.6 & 0.0 & 11.3 & 0.0  \\
\hline
\end{tabular}}
\end{table}

We replace $\hat{Q}\hat{K}\hat{V}s$ with vanilla $QKVs$ and self mask with padding mask, respectively. The padding mask is an attention mask that only masks the attention connections of padded tokens in a mini-batch and is widely used in transformer models \cite{Yosuke2020maskctc,YEBAI2020laji,song2021duokuileni,TSNAT2021TWOPASS}.

After removing $\hat{Q}\hat{K}\hat{V}s$ and self mask all or separately, the accuracies on the dev set approach 100\% quickly, and the training processes are stopped because of early stopping, resulting in few training epochs and poor CTC performance. However, we only focus on the CERR between the greedy CTC output and decoder output. As shown in Table ~\ref{tab:leakage}, except for NAT-UBD, the decoder outputs of the other three models are the same as the greedy CTC output, indicating that these three decoders have learned identity mapping between input and output during training. The experimental phenomena verify that information leakage during training can damage the network performance, proving both the $\hat{Q}\hat{K}\hat{V}s$ and self mask should be used.

\subsection{Effectiveness of Adaptive Termination}

\begin{table}[t]
\caption{Ablation study of adaptive termination on the Aishell1 corpus.}
\label{tab:termination}
\centering
\scalebox{0.95}{
\begin{tabular}{lcccc}
\hline
Model & $J$ & Dev & Test & RTF \\
\hline
NAT-UBD & $10$ & 5.0 & 5.5 & \textbf{0.0116} \\
$-$adatpive termination & 10 & 5.0 & 5.5 &0.0263 \\
\hline
\end{tabular}}
\end{table}
We remove the adaptive termination from NAT-UBD during decoding. From Table ~\ref{tab:termination}, we can conclude that the adaptive termination can improve the decoding speed more than 2$\times$ times with $J=10$ while keeping CERs unchanged.

\subsection{Effectiveness of Unified Bidirectional Decoder}

\begin{table}[t]
\caption{Comparison of the left-to-right (L2R) decoder, right-to-left (R2L) decoder and unified bidirectional decoder (UBD) on the Aishell1 corpus and Magicdata corpus.}
\label{tab:contexts_magicdata}
\centering
\scalebox{0.95}{
\begin{tabular}{cccccccc}
\hline
\multicolumn{1}{c}{\multirow{2}{*}{Decoder}} & {\multirow{2}{*}{$J$}} & \multicolumn{3}{c}{Aishell1} & \multicolumn{3}{c}{Magicdata} \\ \cmidrule(r){3-5} \cmidrule(r){6-8}
\multicolumn{1}{c}{} & & Dev & Test & RTF & Dev & Test & RTF                 \\
\hline
\multirow{2}{*} {L2R}
& 1 & 5.4  & 5.9 & 0.0081 & 4.7 & 5.0 & 0.0095\\
& 10 & 5.5  & 5.9  & 0.0108 & 4.7 & 5.0 & 0.0115 \\
\hline
\multirow{2}{*} {R2L}
& 1 & 5.4  & 5.9 & 0.0081 & 4.7 & 5.0 & 0.0095 \\
& 10 & 5.6  & 6.2  & 0.0103 & 4.7 & 5.1 & 0.0110 \\
\hline
\multirow{2}{*} {L2R+R2L}
& 1 & 5.6  & 6.2 & \textbf{0.0077} & 4.8 & 5.2 & \textbf{0.0090} \\
& 10 & 5.6  & 6.3  & 0.0091 & 4.8 & 5.2 & 0.0101 \\
\hline
\multirow{2}{*} {UBD}
& 1 & 5.1  & 5.6 & 0.0081 & 4.4 & 4.9 & 0.0095 \\
& 10 & \textbf{5.0}  & \textbf{5.5}  & 0.0116 & \textbf{4.2} & \textbf{4.7} & 0.0121 \\
\hline
\end{tabular}}
\end{table}
We verify the effectiveness of UBD by replacing UBD with the L2R decoder and R2L decoder. For fairness, when using L2R and R2L decoders simultaneously, each decoder has 3 decoder layers, and the greedy output with the higher average probability is chosen as the final output. In addition, adaptive termination is used during decoding.

As shown in Table ~\ref{tab:contexts_magicdata}, UBD achieves the best CERs on two corpora with $J=1$. With $J=10$, the CERs of UBD can be further reduced on two corpora while other methods remain unchanged or even worse. In cases with L2R and R2L decoders simultaneously, the decoding speed is faster than other methods due to the parallel decoding of two decoders. However, this method yields worse CERs than only using a unidirectional decoder. It is because the two decoders have no information exchange, which means their outputs only depend on unidirectional contexts. Moreover, both decoders only have 3 decoder layers, which can reduce their respective feature extraction ability. As a result, we can conclude that the proposed UBD is efficient and can use ample linguistic information for character prediction.

However, the RTF of UBD is higher than other methods both on Aishell1 and Magicdata corpora with $J=10$. When using UBD, we observe that the cyclic dependency of adjacent tokens often arises, making adaptive termination invalid and reducing the decoding speed. For example, the ground truth is ``stand up'', and the output of the first iteration is ``stand down''. Then the outputs of the second and third iterations might be ``sit up'' and ``stand down''. As a result, the iteration can not stop until ten iterations. Especially, cyclic dependency arises with increasing frequency when the phonetic pronunciation is similar.

\section{Conclusion}
\label{sec:Conclusion}

We propose a new non-autoregressive transformer with a unified bidirectional decoder (NAT-UBD), carefully designed to simultaneously utilize left-to-right and right-to-left contexts and prevent consequent information leakage.  As a result, the proposed NAT-UBD outperforms all previous NAR transformer models on the Aishell1 corpus and achieves competitive performance with AR transformer on the Magicdata corpus without any external language model. For the decoding speed, NAT-UBD can run 49.8$\times$ faster than the AR transformer baseline. Further analysis experiments prove the effectiveness of the unified bidirectional decoder and the necessity of avoiding information leakage. We plan to reduce exposure error from feeding ground truth to the decoder during training and greedy CTC output during decoding.

\bibliographystyle{IEEEbib}
\bibliography{myref}

\end{document}